\documentclass[runningheads]{llncs}
\usepackage[T1]{fontenc}
\usepackage{graphicx}
\usepackage{subfigure} 
\usepackage{multirow}
\begin{document}
\title{Towards Building a Robust Knowledge Intensive Question Answering Model with Large Language Models}
\titlerunning{Towards Building a Robust Knowledge Intensive QA model with LLMs}
%
\author{Xingyun Hong\inst{1} \and
Yan Shao\inst{1,2,*} \and
Zhilin Wang\inst{1} \and
Manni Duan\inst{1} \and
Xiongnan Jin\inst{1}}

\authorrunning{XY. Hong et al.}

\institute{Zhejiang Lab, China\\
\email{\{xyhong,zlwang, mnduan,xnjin\}@zhejianglab.com}\\
\and
China Mobile, Hangzhou Research and Development Center, China \\
\email{shaoyan@cmhi.chinamobile.com}
}

\maketitle              
\begin{abstract}
The development of LLMs has greatly enhanced the intelligence and fluency of question answering, while the emergence of retrieval enhancement has enabled models to better utilize external information. However, the presence of noise and errors in retrieved information poses challenges to the robustness of LLMs. In this work, to evaluate the model's performance under multiple interferences, we first construct a dataset based on machine reading comprehension datasets simulating various scenarios, including critical information absence, noise, and conflicts. To address the issue of model accuracy decline caused by noisy external information, we propose a data augmentation-based fine-tuning method to enhance LLM's robustness against noise. Additionally, contrastive learning approach is utilized to preserve the model's discrimination capability of external information. We have conducted experiments on both existing LLMs and our approach, the results are evaluated by GPT-4, which indicates that our proposed methods improve model robustness while strengthening the model's discrimination capability.

\end{abstract}

\section{Introduction}
With the development of deep learning and reinforcement learning, large language models (LLMs) have demonstrated significant potential in various natural language processing (NLP) applications\cite{kojima2022large}. As LLMs continue to evolve, concerns regarding the reliability have risen, such as outdated information and fabricated outputs \cite{ji2023survey}. Outdated information stems from static pre-training data, leading to model's lack of the latest knowledge. Moreover, the hallucination of model may mislead its user, undermining the credibility of LLMs. To address these issues, retrieval-augmented generation (RAG) \cite{asai2023retrieval} has been proposed and combined with LLMs. By incorporating external information, LLMs can adapt to dynamic contexts and provide more reliable and relevant results.

\begin{figure}
    \centering
    \includegraphics[width=0.9\columnwidth]{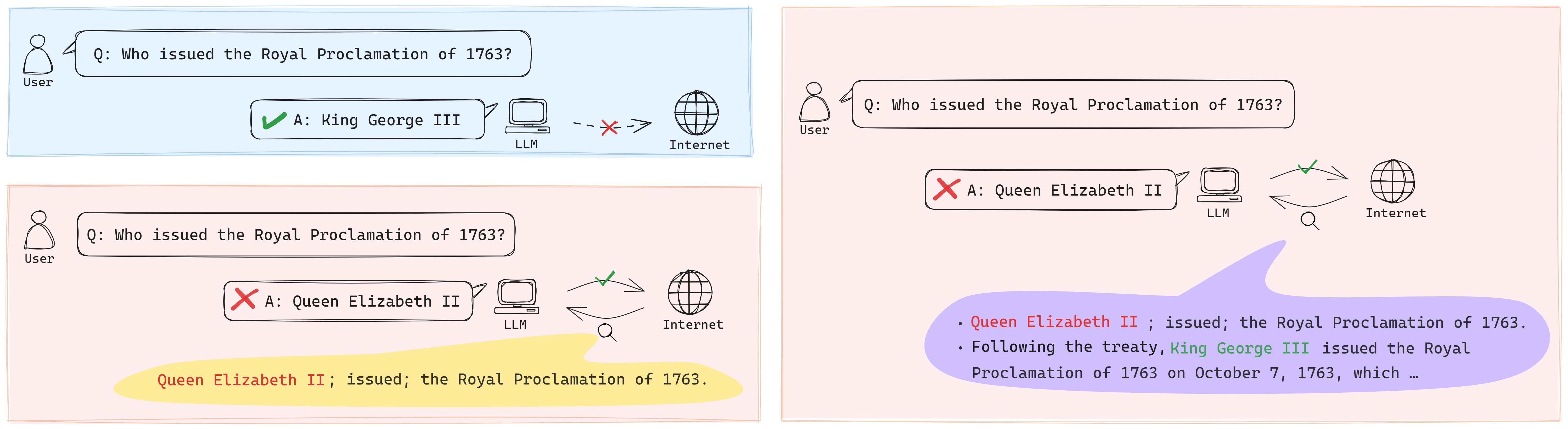}
    \vspace{-3mm}
    \caption{An example of how external information impacts ChatGLM3-6B's performance.}
    \label{fig:example}
\end{figure}

While retrieval-augmented enhancement proven to be helpful, it introduces its own set of challenges. External knowledge sources vary in terms of reliability, structure and quality. For example, knowledge from web search is updated more frequently, its format is relatively free and contains more noise; while knowledge from structured knowledge base like knowledge graph is more static and unified. The diversity of information may result in inconsistencies or conflicts. Figure~\ref{fig:example} demonstrates the influence of external information on ChatGLM3-6B. When there is no external information, the model provides correct answer; however, when incorrect context or conflicting information is presented, the model tends to give incorrect answer. Previous studies also show that adding random or irrelevant context can decrease QA performance \cite{shi2023large}. Therefore, how to enhance LLM's ability of utilizing various information is worth studying.

\vspace{-3mm}
\vspace{-3mm}

Addressing concerns that adding certain question-related noise affects LLM's robustness, most of other work involves filtering of information in preliminary stages or controlling whether LLM should conduct additional retrieval. However, such prefixed processes do not always eliminate irrelevant information. Hence, we focus on strengthening LLM's discrimination ability to enhance its robustness.

In this work, to evaluate the robustness of LLM in retrieval scenarios, we first utilize datasets from machine reading comprehension (MRC) to create new datasets simulating various retrieval scenarios. Additionally, to effectively tackle challenges posed by critical information absence, noise and conflicts, we fine-tune LLM integrating data augmentation techniques including mask and swap to improve its performance on the question-answering task. Furthermore, to reinforce the model's ability to refrain from providing answers beyond its scope rather than generating inaccurate responses, we introduce a contrastive learning-based approach. By constructing training data based on whether the model knows the answer, this approach enhances model robustness while strengthening its discrimination capability.

The contributions of this paper can be summarized as follows:\footnote{https://github.com/sherryhongxy/Training-a-robust-QA-model-with-LLMs}
\begin{itemize}
    \item We introduce a dataset construction method for evaluating model question answering accuracy under different interferences.
    \item We evaluate and analysis the performance of LLMs with various kinds of external information systematically.
    \item A data augmentation-based fine-tuning method for LLM is proposed to enhance its accuracy and robustness against various interferences.
    \item A contrastive learning-based method is proposed to enhance model's discrimination capability and utilization of both external information and its internal knowledge.
\end{itemize}

\section{Related Work}
\subsection{Retrieval-Augmented LLMs}
The role of retrieval enhancement encompasses enhancing inferential capabilities, elevating answer traceability, and alleviating model hallucination. External
information can be obtained from web, knowledge base, database and other sources. \cite{lazaridou2022internet} incorporates knowledge into prompts. \cite{trivedi2022interleaving} utilizes search and chain-of-thought, allowing models to follow logical sequences and retrieve information in a more contextually relevant manner. Some studies also teach LLMs to use external tools including retriever, calculator, and other foundation models \cite{shen2023hugginggpt}. In addition to merely use retriever, \cite{lewis2020retrieval} collaboratively optimizes both retrieval models and language models. 

While retrieval enhancement has demonstrated its efficiency, some studies indicate that the inclusion of irrelevant information can impact model performance. Consequently, the implementation of retrieval enhancement involves considerations such as when to invoke retrieval \cite{wang2023self}, the selection of external evidence \cite{shi2023replug}, and post-evaluation of generated results \cite{asai2023self}. These factors play
crucial roles in fine-tuning the retrieval enhancement process and ensuring that the model leverages external knowledge judiciously to enhance its performance.

\subsection{Robustness of LLMs}
The robustness of LLMs is a crucial factor in application, typically evidenced by their performance under attack or disruptive inputs. Depending on where the perturbations occur, the study of model robustness can be classified into prompt robustness and task robustness. 

PromptBench \cite{zhu2023promptbench} constructs adversarial prompt datasets, perturbing prompts at multiple level to evaluate how slight deviations, such as spelling errors or synonyms, affect LLM results while maintaining semantic integrity. \cite{greshake2023more} is the work related to prompt injection attack. 

Task robustness is to observe model performance by perturbing different tasks such as sentiment analysis, natural language inference, classification and so on \cite{wang2023robustness} with typos, grammatical errors, and insertions. Some datasets are designed to evaluate model robustness, including multi-task benchmark AdvGLUE \cite{wang2021adversarial}, table-based question-answering dataset RobuT \cite{zhao2023robut}, and others focusing on code generation, math reasoning and dialogue generation.

\section{Dataset Construction} 
In practical applications, deploying LLMs for knowledge-based question answering may encounter challenges when retrieving external knowledge through external searches. This is due to the potential presence of irrelevant or erroneous information, as well as variations in format among different knowledge sources. 

\begin{figure*}
    \centering
    \includegraphics[width=0.95\columnwidth]{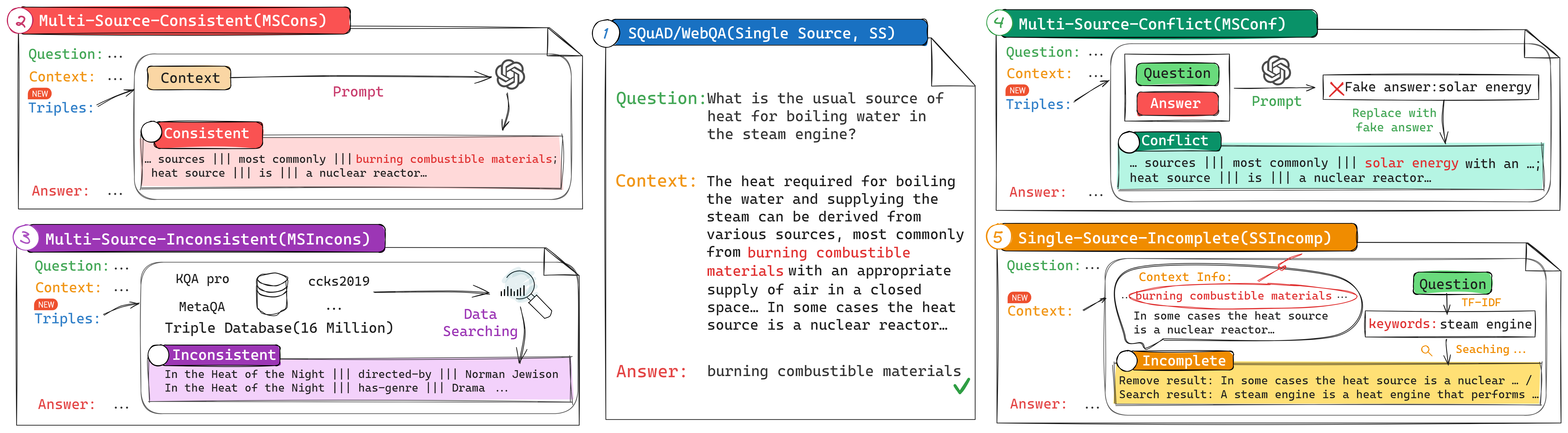}
    \caption{Dataset construction: methods to create five kinds of samples.}
    \label{fig:dataset}
\end{figure*}

To evaluate the performance of LLMs under such circumstances and enable them to better handle diverse scenarios, we first choose two MRC datasets, then we apply several data construct techniques to generate new datasets. The correctness of the generated samples is evaluated by rule-based methods and human review. According to the context type, the new constructed samples are categorized into five classes. Figure \ref{fig:dataset} illustrates the process of dataset construction.

\vspace{-3mm}
\vspace{-3mm}


\textbf{Single Source (SS)}
The selection of MRC datasets serves various purposes in our work. These datasets typically comprise questions paired with corresponding contexts, from which the answers are extracted. This structure facilitates the addition of noise and the replacement of answers. Additionally, the datasets encompass a significant portion of questions that rely on common sense, which LLMs likely to have encountered during pre-training. As a result, these datasets allow us to examine models' discrimination ability under the impact of misleading external information.

SQuAD (in English) \cite{rajpurkar2016squad} and WebQA (in Chinese) \cite{li2016dataset} are chosen as our base datasets. SQuAD is a MRC dataset consisting of 100k+ questions posed by crowd workers on \ Wikipedia articles, where the answer to each question is a segment of text from the corresponding passage. WebQA is a large scale human annotated real-world QA dataset with more than 42k questions and 556k evidences.  

We sample 500 instances from each dataset to form a  development set and a test set respectively. The proportion of development and test set is 1:1. The original data contains question, question-related context and a corresponding answer. This dataset is referred as \textbf{SS} below. 

\textbf{Single-Source-Incomplete (SSIncomp)}
Incomplete text is designed for assessing LLMs' performance with relevant but insufficient data. In this case, the topic of context is partly relevant to the question, but the crucial information is absent from the context. Traditional MRC model can not handle such situation. As for the LLM, endowed with internal knowledge, is anticipated to determine whether the provided context sufficiently addresses the question, and leverage its internal knowledge to generate appropriate responses if the answer can not be inferred from the context.

In SQuAD, context originates from article paragraphs, our approach involves the removal of sentences containing answers, while retaining the remainder. However, context in WebQA is typically short and answer-centric, making the elimination process yield few information. Therefore, we employ TF-IDF to identify keywords in the question, leveraging them for online search.\footnote{We utilize the SerpAPI service for Google search: https://serpapi.com/.} Subsequently, the search result that devoid of the answer is retained as context.

\textbf{Multi-Source-Consistent (MSCons)}
When retrieving information, data from different sources may exhibit diverse formats and content variations. Our objective is to explore impact of multi-source information on model performance. Specifically, we aim to observe whether information with similar content but different formats affects the model's inference results. Since the original data is in natural language, we utilize GPT3.5-Turbo \cite{ouyang2022training} to extract multiple sets of triples as an alternative data source. We retain those samples for which answers exist in the extracted triples. 

\textbf{Multi-Source-Inconsistent (MSIncons)}
In addition to examining scenarios with consistent content across multiple sources, we anticipate the model to remain robust even if the retrieved results contain additional noise, which may be partially relevant or irrelevant to the question. Therefore, we build a triple database using several datasets, for instance, KQA pro, MetaQA dataset and so on, containing 163,776,434 triples in total. These triples are not necessarily related to the questions in WebQA and SQuAD. We utilize question or the head entities extracted by GPT3.5-Turbo as search terms, and ensure that the answers are not included in the retrieved results. The final selected triples are limited to 10 with respect to relevance. 

\textbf{Multi-Source-Conflict (MSConf)}
Given the substantial variability in the quality of internet information, retrieval process occasionally yields erroneous or conflicting results. As such situations are often inevitable, it becomes imperative for LLMs to discriminate between correct and incorrect information when presented simultaneously. In this case, we have devised conflicting samples. First, we use GPT3.5-Turbo to generate a similar yet incorrect answer based on the question and original answer. Subsequently, the answer in the triples consistent with the context is substituted with the fabricated one.  

\textbf{Dataset Quality}
Given the potential errors introduced by construction methods and tools, during dataset construction, we conduct multiple case-based optimizations of the prompt. For created samples, the unsatisfactory results are filtered out from our test set, ensuring the generated results align with our requirements. Specifically, for SSIncomp and MSIncons samples, we ensure that answers are not included in the deleted or retrieved results. For MSCons samples, triplets generated by GPT3.5-Turbo must contain corresponding answers. For MSConf samples, 100 instances are randomly sampled for human review. The plausibility of the generated false answers, considering the correspondences of the answer type (such as names, numbers, locations, times and so on), achieving an accuracy rate of 99\%, guaranteeing that the generated samples meet the experimental requirements.

\vspace{-3mm}
\begin{figure*}
    \centering
    \includegraphics[width=0.9\columnwidth]{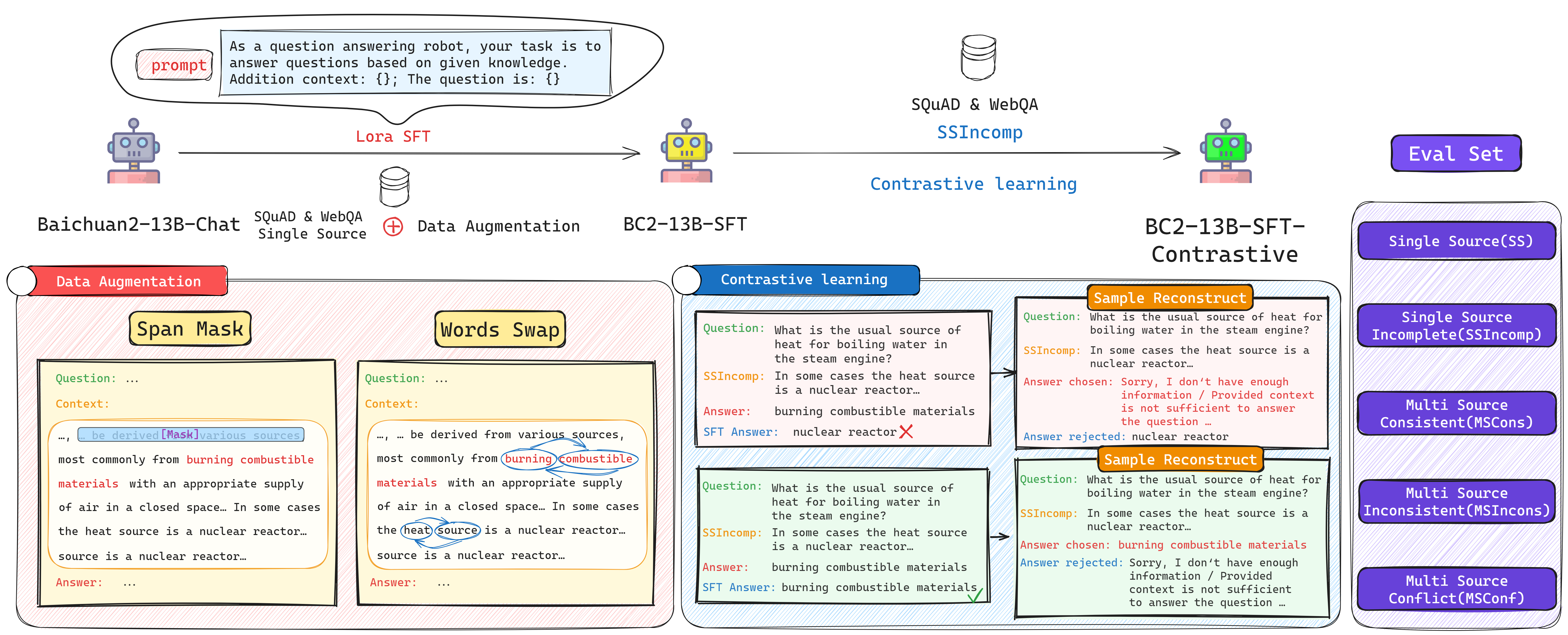}
    \caption{Two-stage fine-tuning with the Baichuan2-13B-Chat model}
    \label{fig:process}
\end{figure*}
\vspace{-7mm}

\section{Methods}

\subsection{Data Augmentation}
To enhance the model's robustness under diverse contexts, we integrate data augmentation techniques during model fine-tuning. Original data in the SQuAD and WebQA dataset contains questions, context and answers. Typically, the answer is contained in the question-related context. Leveraging corresponding and comprehensive context to the question facilitates the task for LLMs, yet the presence of noise within the context may weaken the model's generalization ability. Consequently, we implement span masking and word swapping strategies to enhance the model's robustness.

\textbf{Mask}: 
We enhance the model's reasoning and generalization abilities by masking certain portions of the context. Even if the context does not explicitly contain the answer, LLM can leverage context related to the answer and its internal knowledge to improve accuracy. Masking is performed with spans, typically short sentences whose removal do not significantly affect semantic integrity.

Every span (usually delimited by two punctuation marks, using regex to separate) has equal possibility of being eliminated. The absence of answer motivates LLM to distinguish the relevance of information and stick to its own knowledge when confronted with inadequate context, whereas the removal of other span simulates the incompleteness of information.

\textbf{Swap}: As disorder of words between adjacent words does not alter its overall meaning significantly, the switch of words can be regarded as adding noise to the context. In our approach, we randomly select a span from the context and a position within span, then interchange adjacent word sets.

Unlike methods involving open-ended searches or GPT generation, these approaches have comparatively low costs in constructing training data, thus are more practical for real-world applications.

\subsection{Contrastive Learning} 
Through detailed case analysis (refer to Sec5.2), we discover that LLMs possess different levels of discrimination capabilities. When confronted with queries that cannot be answered with the given information, the model occasionally informs the user of the inadequacy of the provided knowledge. To strengthen model's discrimination ability, we employ contrastive learning after first stage fine-tuning.

As to find what does the model know and guide it to decline to the question beyond its knowledge, we create SSIncomp samples of each dataset as training data. For SQuAD, elimination of answer-contained sentence is applied. For WebQA, context relevant to question but does not include the answer is selected. 

First, the fine-tuned model generates outputs for the aforementioned samples, subsequently assessed by GPT-4. To facilitate contrastive learning, we construct new dataset. Each question and context pair is accompanied by both an answer to accept and one to decline, as depicted in Figure ~\ref{fig:process}.

For correctly-inferred samples, the original sample label is designated as the chosen answer, while a specific expression is assigned as the rejected answer, such as "Provided context is not sufficient to answer the question/Sorry, I don't have enough information...". As for incorrect samples, the sentence indicating model's inability to respond constitutes the chosen answer, whereas the incorrect output from the fine-tuned model serves as the declined answer.

Throughout the training process, we aim to maximize the disparity in probability between accepted and declined answers through comparison.

Specifically, the loss function is defined as:
\begin{small}
\begin{equation}
L = -\sum_{i=1}^{N}{log\sigma(\frac{1}{C}{\sum_{i=1}^{C}logp(y_{C_i}|x)}-\frac{1}{R}\sum_{i=1}^{R}logp(y_{R_i}|x)})
\end{equation}
\end{small}
\noindent where $N$ represents the sample numbers, $\sigma$ is the sigmoid function. $C$ and $R$ stands for tokens of chosen label and rejected label respectively. $logp(y_{C_i}|x)$ is the log probability of the chosen token $C_i$ when given $x$ as input.

\section{Experiment}
\subsection{Models}
To examine the impact of various information on LLMs, we initially evaluate model performance using the dataset we construct. Subsequently, we select a base model to conduct further experiments. Below are the models utilized for our experimental evaluation.

\textbf{GPT3.5-Turbo}: GPT3.5-Turbo is a conversational artificial intelligence model developed by OpenAI, based on the GPT architecture. 

\textbf{Baichuan}\cite{yang2023baichuan}: Generation of LLM from Baichuan Intelligence and supports both Chinese and English. 

\textbf{Llama}\cite{touvron2023llama}: A large language model released by Meta AI, which supports text completion and chat completion. 

\textbf{ChatGLM}\cite{du2021glm}: 
ChatGLM is an open bilingual language model (supports English and Chinese) based on General Language Model (GLM) framework.

\subsection{Evaluation Metrics}
Considering that the outputs of the original LLM are typically more comprehensive and long, whereas the labels in the MRC dataset are comparatively concise, utilizing n-gram metrics such as ROUGE and BLEU can not indicate real performance precisely. Therefore, for each sample, we employ recall and accuracy to evaluate. \textit{Recall} assesses the percentage of overlapping words between the model outputs and labels, it's suitable for comparing model performance after fine-tuning, since the output format remains consistent.

While recall focuses on words matching, we utilize GPT-4 to determine whether the whole output aligns the key points of the label, even if in different ways of expression. In practical application, when querying LLM, it's preferable to receive a response indicating model's inability to answer rather than an incorrect one. Considering this scenario, GPT-4 is used to categorize the model's inferences into wrong (\textit{w}), correct (\textit{c}), or rejected (\textit{r}, indicating the model declines to give answer directly due to lack of information or other reasons). 

The standard accuracy metric (\textit{ACC}) is defined as the proportion of the correct samples among all samples. Meanwhile, to further distinguish the model's ability to decline questions beyond its ability, correct, rejected and incorrect responses receive a score of 1, 0, and -1 respectively. The average score is then computed across all samples as the weighted score (\textit{WSCORE}).

\subsection{Experimental Setup}
In the experiment, we commence by assessing the performance of several existing LLMs using our dataset. Among the open-source models, Baichuan2-13B-Chat demonstrates satisfactory performance, coupled with a relatively swift inference speed, thus making it a suitable candidate for further experiments.

Due to the resource limitation and the effectiveness of LoRA\cite{hu2021lora}, we fine-tune LLMs based on LoRA framework. The rank of the adaptors is set as 8, maximum input length is 2048, and the learning rate is 1e-4 with a warm-up strategy. The experiment is accelerated using NVIDIA V100 GPUs with 32GB memory each.

During the first fine-tuning stage, approximately 40\% of answer-located span are masked. In the second fine-tuning stage, we select 3,500 new samples, ensuring a balanced distribution of positive and negative instances at 1:1.

\subsection{Results}
\textbf{Existing model performance evaluation} 
As shown in Table \ref{tab:model-eval}, we observe that the accuracy of nearly every model (excluding Llama2-7B-Chat) surpasses 85\% on the original single-source samples (SS). In this case, the context aligns perfectly with the question, apart from a few inference-based questions, which are challenging to answer.

However, for samples where the answer is absent in the context (SSIncomp), there is a notable decline in accuracy. This discrepancy can be attributed to certain questions heavily relying on contextual cues. To isolate this effect, we evaluate the accuracy solely based on the question using Baichuan2-13B-Chat. The accuracy is 52.2\%, indicating that approximately half of the questions can be answered based on the internal knowledge of the Baichuan2-13B-Chat model alone. This figure surpasses Baichuan2-13B-Chat's accuracy on SSIncomp samples (48.8\%), underscoring the disruptive impact of incomplete or irrelevant contextual information on the model's performance.

\vspace{-5mm}
\begin{table*}[ht!]\scriptsize
\centering
\caption{Evaluation of different models on the test set: \textit{ACC} stands for accuracy, \textit{R} stands for rejection percentage, \textit{WSCORE} is the weighted accuracy. Model name with * indicates it is a commercial model.}
\begin{tabular}{|l|c|c|c|c|c|c|c|c|c|c|c|c|}
\hline
\textbf{Model} & \multicolumn{2}{c|}{\textbf{SS}} &\multicolumn{2}{c|}{\textbf{SSIncomp}} &\multicolumn{2}{c|}{\textbf{MSCons}} &\multicolumn{2}{c|}{\textbf{MSIncons}} &\multicolumn{2}{c|}{\textbf{MSConf}} & \multicolumn{2}{c|}{\textbf{Overall}}\\ 
\hline
\textbf{Metrics} & {ACC}&{R} & {ACC}&{R} & {ACC}&{R} & {ACC}&{R} & {ACC}&{R} & {ACC}&{WSCORE}\\ 
\hline
GPT3.5-Turbo* & 96.7 & 0.0 & 60.0 & 3.3 & 95.9 & 0.6 & 93.9 & 1.6 & 72.4 & 0.8 & 83.8 & 68.8 \\
\hline
Llama2-7B-Chat & 78.5 & 0.2 & 34.6 & 8.1 & 81.3 & 0.0 & 74.0 & 0.0 & 60.2 & 0.0 & 65.7 & 33.1\\
Llama2-13B-Chat & 86.2 & 1.2 & 38.2 & 8.7 & 90.0 & 0.4 & 85.2 & 0.0 & 65.9 & 0.2 & 73.1 & 48.3\\
Baichuan2-7B-Chat & 92.7 & 0.6 & 49.8 & 6.5 & 91.3 & 0.8 & 90.9 & 0.6 & 70.1 & 1.0 & 79.0 & 59.8 \\
Baichuan2-13B-Chat & \textbf{93.9} & 0.4 & 48.8 & 20.5 & 90.2 & 0.6 & 88.2 & 3.5 & 70.7 & 0.8 & 78.3 & 61.9\\
ChatGLM2-6B & 85.8 & 1.2 & 33.3 & 15.4 & 85.4 & 0.6 & 83.9 & 1.4 & 75.4 & 0.2 & 72.8 & 49.3\\
ChatGLM3-6B & 91.1 & 0.2 & 39.6 & 13.0 & 89.8 & 0.6 & 87.0 & 0.2 & 75.6 & 0.8 & 76.6 & 56.2\\
\hline
BC2-13B-SFT & 93.7 & 0.0 & \textbf{55.5} & 0.0 & \textbf{95.3} & 0.0 & \textbf{93.7} & 0.0 & 77.8 & 0.0 & \textbf{83.2} & 66.4 \\
BC2-13B-SFT-Contrastive & 92.7 & 2.8 & 48.4 & 21.3 & 93.5 & 1.6 & 92.9 & 2.0 & \textbf{78.5} & 2.8 & 81.2 & \textbf{68.5} \\
\hline
\end{tabular}
\label{tab:model-eval}
\end{table*}
\vspace{-5mm}

In scenarios involving multiple sources, even though the information extracted from triples corresponds to the context and contains answers (MSCons samples), its variability due to extraction still impacts model's robustness. While most models experience a slight decrease in accuracy, Llama models show improved performance, suggesting the proficiency in handling formally structured knowledge. 
If triples are retrieved from triple database (MSIncons samples), the accuracy further decreases. In such cases, the model becomes susceptible to unrelated triples, disregarding correct contextual information.

Finally, when we construct false answers which is similar to real ones and replace the information in the triples (MSConf samples), the model easily falls into hallucinations when faced with two conflicting contexts. Consequently, erroneous answers are generated, highlighting the vulnerability of the model in distinguishing between contradictory information.

From the perspective of how different model handles the situation of noise, GPT3.5-Turbo outperforms others in every scenario except when confronted with conflicting contexts (MSConf). Despite ChatGLM's general inferiority to other models, its capability of managing conflict is surprisingly noteworthy. 

\textbf{Fine-tuning model performance evaluation}
BC2-13B-SFT refers to the Baichuan2-13B-Chat model after fine-tuning utilizing data-augmentation strategies. As demonstrated in Table \ref{tab:model-eval}, compared to the original Baichuan2-13B-Chat, there are significant improvements in almost every scenarios. The overall accuracy increase by 6.3\%, only slightly lower than GPT3.5-Turbo. Since GPT3.5-Turbo is a closed source model, the fine-tuned model offers the advantage of being more suitable for localized deployment.


The adopted training data is merely in single source format , however it still contributes to the improvement in every situation. This suggests that related but incomplete context motivates LLM to leverage its internal knowledge and thereby promoting its discrimination ability.

\textbf{Evaluation of model's discrimination ability}
We further distinguish declined responses from incorrect responses to observe the proportion of rejection on SSIncomp and MSConf samples. Meanwhile, in other three scenarios, since the context contains the answer, though potentially with noise, we still anticipate correct responses from the model.

The metric \textit{r} represents the proportion of declined responses. Each model demonstrates varying levels of discrimination capability, with Baichuan2-13B-Chat and  ChatGLM exhibiting relatively strong discrimination abilities. However, a higher rejection rate implies less informative responses. So it is necessary to assess this metric in conjunction with the correct answer proportion. In terms of the overall \textit{WSCORE}, GPT3.5-Turbo remains the best. Despite its lower rejection rate, the proportion of correct answers significantly surpasses other models, followed by the Baichuan2-13B-Chat model.

As for the BC2-13B-SFT model, after fine-tuning with MRC samples, the model tends to provide answers even if they are incorrect. However, after second stage fine-tuning utilizing contrastive learning, model's discrimination capability is enhanced, transitioning from delivering an answer (regardless of correctness) to declining to the question beyond its capacity. Moreover, in the scenarios involving multi-source, the accuracy remains superior to that of original Baichuan2, maintaining robustness against noise. There is an increment of 10.7\% in the overall weighted score of BC2-13B-SFT-Contrast compared to the original Baichuan2-13B-Chat model and approaching the performance of GPT3.5-Turbo. It is worth noting that despite the relatively small number of training samples, contrastive learning contributes to improving the model's discrimination capability.

\vspace{-5mm}
\begin{table*}[ht!]\scriptsize
\centering
\caption{Case study: The Mitchell Tower is designed to look like what Oxford tower? (Note the responses from some models are reduced to keywords for display purposes)}
\begin{tabular}{|p{1cm}|p{5.5cm}|p{5.5cm}|}
\hline
Sample & Additional context &  Model response \\
\hline
\multirow{4}{1cm}{SS} & \multirow{4}{5.5cm}{Text: ...Mitchell Tower, for example, is modeled after Oxford's \textbf{Magdalen Tower}, and the university Commons, Hutchinson Hall, replicates Christ Church Hall.} & GPT3.5-Turbo: magdalen tower \\
~ & ~ & Baichuan2-13B-Chat: magdalen tower \\
~ & ~ & BC2-13B-SFT: magdalen tower \\
~ & ~ & BC2-13B-SFT-Contrast: magdalen tower \\
\hline
\multirow{4}{1cm}{SS-Incomp} & \multirow{4}{5.5cm}{Text:The first buildings of the University of Chicago campus, which make up what is now known as the Main Quadrangles, ..., patterned on the colleges of the University of Oxford.} & GPT3.5-Turbo: magdalen tower \\
~ & ~ & Baichuan2-13B-Chat: it is not meant to represent a specific tower\\
~ & ~ & BC2-13B-SFT: bodleian library \\
~ & ~ & BC2-13B-SFT-Contrast: sorry, I can't answer. \\
\hline
\multirow{4}{1cm}{MS-Cons} & \multirow{4}{5.5cm}{Text:...Mitchell Tower, for example, is modeled after Oxford's \textbf{Magdalen Tower}... Triples: ..., Mitchell Tower ||| modeled after ||| Oxford's \textbf{Magdalen Tower}, university Commons, ...} & GPT3.5-Turbo: magdalen tower \\
~ & ~ & Baichuan2-13B-Chat: magdalen tower \\
~ & ~ & BC2-13B-SFT: magdalen tower \\
~ & ~ & BC2-13B-SFT-Contrast: magdalen tower \\
~ & ~ & \\
\hline
\multirow{4}{1cm}{MS-Incons} & \multirow{4}{5.5cm}{Text:...Mitchell Tower, for example, is modeled after Oxford's \textbf{Magdalen Tower}... Triples: The Deadly Tower ||| directed by ||| Jerry Jameson, ...} & GPT3.5-Turbo: magdalen tower \\
~ & ~ & Baichuan2-13B-Chat: magdalen tower \\
~ & ~ & BC2-13B-SFT: magdalen tower \\
~ & ~ & BC2-13B-SFT-Contrast: magdalen tower \\
\hline
\multirow{4}{1cm}{MS-Conf} & \multirow{4}{5.5cm}{Text:...Mitchell Tower, for example, is modeled after Oxford's \textbf{Magdalen Tower}... Triples: ... Mitchell Tower ||| modeled after ||| Oxford's \textbf{Radcliffe Camera}, university Commons, ...} & GPT3.5-Turbo: oxford's Radcliffe Camera \\
~ & ~ & Baichuan2-13B-Chat: oxford's Radcliffe Camera \\
~ & ~ & BC2-13B-SFT: magdalen tower \\
~ & ~ & BC2-13B-SFT-Contrast: magdalen tower \\
\hline
\end{tabular}
\label{tab:case}
\end{table*}
\vspace{-5mm}

\textbf{Case study}
To give a more intuitive visualization, some qualitative comparison results are shown in Table \ref{tab:case}. It is observed that for SS, MSCons and MSIncons samples, the listed models provide correct answers consistently. However, for SSIncomp sample, while GPT3.5-Turbo is capable of generating a correct response based on its internal knowledge, Baichuan2-13B-Chat fails to produce a specific answer. Though fine-tuning reinforces model's confidence on the incorrect answers, BC2-13B-SFT-Contrast, guided by further contrastive learning, appropriately declines to answer when lacking internal knowledge. As for MSConf sample, both Baichuan2-13B-Chat and GPT3.5-Turbo are misled by incorrect triples, whereas fine-tuned models exhibits robustness against such interference.

\subsection{Ablation Study}



\subsubsection{Ablation Study of the First Stage Fine-tuning}
To validate the efficacy of data augmentation strategies, we conduct an ablation study using the dev set.

\begin{itemize}
    \item \textbf{BC2-13B-SFT}: use both mask and swap strategies to process original data.
    \item \textbf{BC2-13B-SFT W/O SWAP}: use only mask to process original data.
    \item \textbf{BC2-13B-SFT W/O MASK \& SWAP}: use original training data without any data augmentation strategy. 
\end{itemize}

\vspace{-7mm}
\begin{table*}\scriptsize
\centering
\caption{Ablation study result on Baichuan2-Chat models.}
\begin{tabular}{|l|c|c|c|c|c|c|}
\hline
\textbf{Model ACC} & \textbf{SS} & \textbf{SSIncomp} & \textbf{MSCons} &\textbf{MSIncons} & \textbf{MSConf} & \textbf{Overall} \\
\hline
Baichuan2-13B-Chat & 93.7 & 46.9 & 91.3 & 87.6 & 67.7 & 77.4\\
BC2-13B-SFT & \textbf{96.9} & \textbf{56.5} & \textbf{98.4} & \textbf{95.9} & \textbf{79.1} & \textbf{85.4}\\
BC2-13B-SFT W/O SWAP & 95.7 & 55.1 & 96.3 & 95.7 & 76.4 & 83.8\\
BC2-13B-SFT W/O MASK \& SWAP & 96.1 & 50.2 & 96.9 & 94.5 & 75.8 & 82.7\\
\hline
\end{tabular}
\label{tab:sft}
\end{table*}
\vspace{-5mm}

First, when compare BC2-13B-SFT with BC2-13B-SFT W/O SWAP, there is a notable improvement in MSConf samples, indicating the disorder of words motivates model's discrimination ability, facilitating more effective selection of the correct answer when faced with conflicting information from two sources. 

Second, when we further eliminate mask strategy in training samples, the performance of SSIncomp samples decreases. This decrease can be attributed to the model's training solely on samples exclusively containing context corresponding to the question, reinforcing the model's reliance on user-provided information, thereby diminishing its discrimination capabilities. As a result, when the provided text lacks essential information, the accuracy of the model decreases.

Finally, training with merely MRC dataset still improve the performance compared to the original Baichuan2 model, which demonstrates that the efficacy of aligning with the corresponding task format.

\subsubsection{Effectiveness of Contrastive Learning}
To validate the effectiveness of contrastive learning, we also compare the preference learning based on direct preference optimization (DPO)\cite{rafailov2024direct} and conventional fine-tuning methods.

In DPO, we utilize the same training samples as in the contrastive learning approach. While DPO yields further improvements in the model's accuracy in multi-source scenarios, it exhibits a relatively higher error rate with SSIncomp samples. This suggests a tendency of the model to provide answers indiscriminately, irrespective of their correctness.

Regarding conventional fine-tuning, the sample's label is determined by either retaining the original label or designating it as a rejected response (refer to sec 4.4), depends on whether the model accurately answers the question. While this fine-tuning approach effectively reduces the occurrence of erroneous outputs on SSIncomp samples, it concurrently impacts the model's discrimination ability in other scenarios severely. Consequently, a substantial portion of responses that could have been deemed correct are instead transformed into rejected outputs.

\vspace{-5mm}
\begin{table*}\scriptsize
\centering
\caption{Evaluation of model's discrimination ability}
\resizebox{\columnwidth}{!}{
\begin{tabular}{|l|c|c|c|c|c|c|c|c|c|c|c|c|c|c|c|c|}
\hline
\textbf{Model} & \multicolumn{3}{c|}{\textbf{SS}} &\multicolumn{3}{c|}{\textbf{SSIncomp}} &\multicolumn{3}{c|}{\textbf{MSCons}} &\multicolumn{3}{c|}{\textbf{MSIncons}} &\multicolumn{3}{c|}{\textbf{MSConf}} & \textbf{Overall}\\ 
\hline
\textbf{Metrics} & {W}&{C}&{R} & {W}&{C}&{R} & {W}&{C}&{R} & {W}&{C}&{R} & {W}&{C}&{R} & WSCORE\\   
\hline
Baichuan2-13B-Chat & 3.7 & 93.9 & 2.4 & 29.3 & 46.9 & 23.8 & 7.7 & 91.3 & 1.0 & 9.6 & 87.6 & 2.8 & 31.5 & 67.7 & 0.8 & 61.1\\
BC2-13B-SFT-Contrast & 3.3 & 93.9 & 2.8 & 30.9 & 44.7 & 24.4 & 2.4 & 96.1 & 1.6 & 3.5 & 93.7 & 2.8 & 18.9 & 78.3 & 2.8 & \textbf{69.5}\\
BC2-13B-SFT-DPO & 3.3 & 96.5 & 0.2 & 46.3 & 46.5 & 7.3 & 3.5 & 96.5 & 0.0 & 5.7 & 94.3 & 0.0 & 20.7 & 79.8 & 0.0 & 66.7\\
BC2-13B-SFT-SFT & 8.3 & 68.7 & 23.0 & 16.5 & 32.9 & 50.6 & 7.9 & 76.4 & 15.8 & 9.8 & 67.5 & 22.6 & 22.4 & 59.7 & 17.9 & 48.0\\
\hline
\end{tabular}
}
\label{tab:res}
\end{table*}
\vspace{-10mm}

\section{Conclusions}
In this paper, we first create dataset simulating various scenarios, including critical information absence, noise, and conflicts, based on MRC datasets to assess various model's performance under multiple interferences. To mitigate the decline in model accuracy attributed to noise, we introduce a data augmentation-based fine-tuning method to enhance LLM's robustness against noise. Additionally, we employ contrastive learning to strengthen the model's discrimination capability. Experimental results indicate that our proposed methods improve model robustness while strengthening the model's discrimination capability.

\begin{credits}
\subsubsection{\ackname} This study was funded by  Independent Project of Zhejiang Lab (K2023NB0AC13), National Natural Science Foundation of China (Grant No.62306287), Zhejiang Provincial Natural Science Foundation of China (Grant No.LY23F020012), Joint R\&D Project of Smart Home Intelligent Interaction Laboratory (R2411A77)
\end{credits}

%
%
%
%

\end{document}